\documentclass[10pt,twocolumn,letterpaper]{article}

\usepackage{wacv}              

\usepackage{graphicx}
\usepackage{amsmath}
\usepackage{amssymb}
\usepackage{booktabs}
\usepackage{multicol}
\usepackage{multirow}
\usepackage{comment}
\usepackage{balance}

\usepackage[pagebackref,breaklinks,colorlinks]{hyperref}

\usepackage[capitalize]{cleveref}
\crefname{section}{Sec.}{Secs.}
\Crefname{section}{Section}{Sections}
\Crefname{table}{Table}{Tables}
\crefname{table}{Tab.}{Tabs.}

\def\wacvPaperID{*****} 
\def\confName{WACV}
\def\confYear{2025}

\begin{document}

\title{Unified Face Matching and Physical-Digital Spoofing Attack Detection}

\author{Arun Kunwar and Ajita Rattani\\
Dept. of Computer Science and Engineering,\\ University of North Texas at Denton, 
USA\\
{\tt\small arunkunwar@my.unt.edu, ajita.rattani@unt.edu}
}
\maketitle

\begin{abstract}   
Face recognition technology has dramatically transformed the landscape of security, surveillance, and authentication systems, offering a user-friendly and non-invasive biometric solution.
However, despite its significant advantages, face recognition systems face increasing threats from physical and digital spoofing attacks.
Current research typically treats face recognition and attack detection as distinct classification challenges. This approach necessitates the implementation of separate models for each task, leading to considerable computational complexity, particularly on devices with limited resources.
Such inefficiencies can stifle scalability and hinder performance.
In response to these challenges, this paper introduces an innovative unified model designed for face recognition and detection of physical and digital attacks.
By leveraging the advanced Swin Transformer backbone and incorporating HiLo attention in a convolutional neural network framework, we address unified face recognition and spoof attack detection more effectively.
Moreover, we introduce augmentation techniques that replicate the traits of physical and digital spoofing cues, significantly enhancing our model robustness.
Through comprehensive experimental evaluation across various datasets, we showcase the effectiveness of our model in unified face recognition and spoof detection.
Additionally, we confirm its resilience against unseen physical and digital spoofing attacks, underscoring its potential for real-world applications.
\end{abstract}

\begin{figure*}
    \centering
    \includegraphics[width=.75\textwidth]{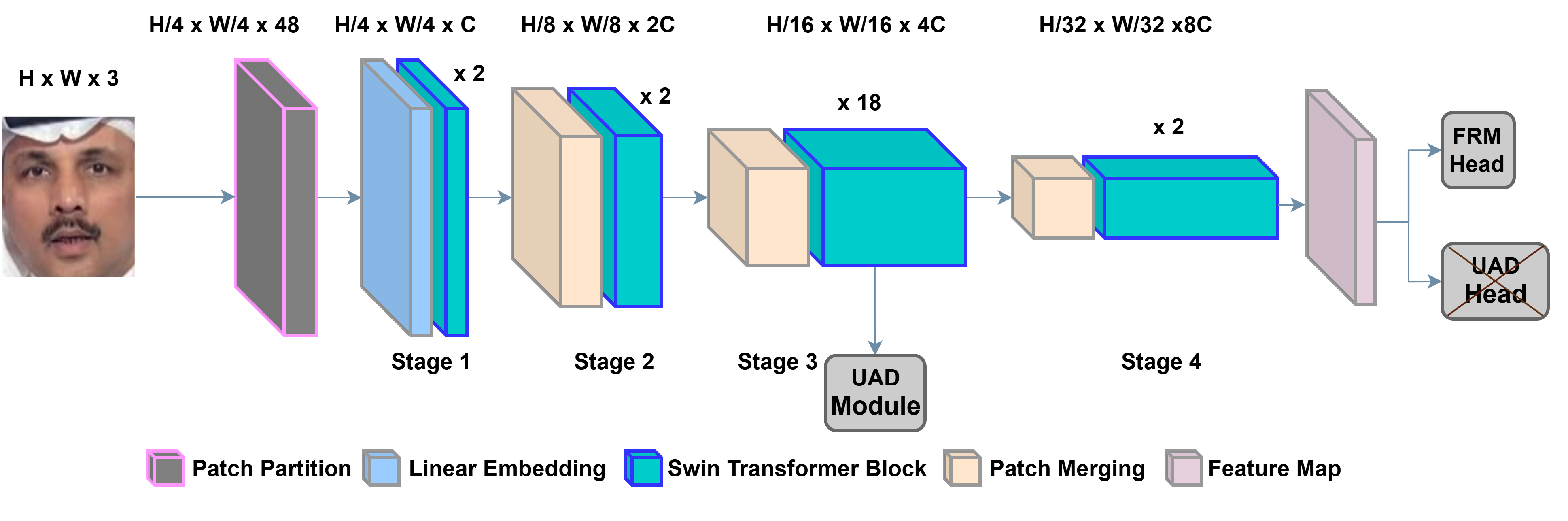}
    \caption{Unified architecture illustrating the integration of the FRM head at the output of the Swin Transformer backbone for face recognition and the UAD module appended at the intermediate layers of Stage 3, specifically for joint physical-digital attack detection. The crossed-out UAD head at the final stage highlights the ineffectiveness of positioning the UAD head at the output of the Swin Transformer backbone. This is because the final layer classification features from the Swin Transformer backbone are not optimal for joint physical and digital attack detection.}
    \label{fig:architecture}
\end{figure*}

\section{Introduction}
\label{sec:intro}

Facial recognition has transformed security and convenience across various sectors, including smartphone unlocking, law enforcement, surveillance, retail personalization, and healthcare, highlighting its wide-reaching impact and transformative potential~\cite{jain2011handbook, parmar2014face}.
It typically operates through a three-stage pipeline: face detection, feature extraction, and face matching.
The detection stage identifies facial regions in an image or video, while the feature extraction stage encodes distinctive facial attributes into high-dimensional vectors using deep learning architectures like Convolutional neural networks~(CNNs) or Transformers. In the matching stage, these feature vectors are compared using distance metrics such as cosine similarity or euclidean distance to verify or recognize identities~\cite{ranjan2018deep,wang2021deep}.

The transformer architecture, with its ability to model global dependencies and spatial relationships through self-attention mechanisms, has obtained state-of-the-art results in image classification, object detection, and segmentation tasks, exceeding traditional CNN-based approaches~\cite{pereira2024review}.
In the context of face recognition, Transformers, such as the Swin Transformer~\cite{liu2021swin}, resulted in the extraction of more discriminative facial features, utilizing self-attention mechanisms, positional encodings, and hierarchical structures to integrate both local and global features.
Thus, achieving state-of-the-art performance in face recognition by addressing challenges such as pose variations, lighting changes, and occlusion~\cite{pereira2024review}.

However, facial recognition systems\footnote{The term face recognition and face matching have been used interchangeably in this paper.} are increasingly vulnerable to adversarial attacks.
Driven by objectives such as financial gain or personal disputes, malicious users often exploit these vulnerabilities by presenting pre-captured or digitally manipulated photos or videos or by using masks and accessories to circumvent the system.
This highlights the critical importance of implementing robust security measures to detect and prevent such threats.
Facial attacks can be broadly categorized into two main categories: \textit{physical} and \textit{digital} spoofing attacks. Physical spoofing attacks involve real-world manipulations like print ~\cite{liu2019multi,zhang2019dataset,zhang2020casia}, replay~\cite{liu2021cross, liu2021casia}, and 3D mask attacks~\cite{fang2023surveillance, jia2020survey,liu20213d,liu2022contrastive }.
Digital attacks include any kind of digital manipulation to the images such as adding  adversarial noise perturbation, image morphing and deepfakes.
Digital spoofing attacks based on facial manipulation based deepfakes depict human subjects with altered identities (identity swap)~\cite{rosberg2023facedancer}, attributes, or malicious actions and expressions (face reenactment) in a given image or a video ~\cite{he2019attgan,choi2018stargan}. Within the \textit{scope} of this study, only digital attacks involving facial manipulations are considered.
Further, they also include attacks based on adversarial noise perturbations in images to mislead the system~\cite{huang2017adversarial,carlini2017adversarial}.

Extensive research has been conducted on facial recognition \cite{jain2011handbook,wang2021deep,zhong2021face}, physical \cite{liu2019multi,zhang2019dataset,zhang2020casia, liu2021cross}, and digital spoofing attack detection~\cite{pei2024deepfake} as distinct classification tasks, each demonstrating significant performance improvement. 
However, deploying separate models for facial recognition, and physical and digital attack detection introduces substantial computational overhead, particularly on resource-constrained devices, thereby limiting efficiency and scalability.
This challenge underscores the need for an integrated (unified) framework that can concurrently address face recognition, with joint physical-digital spoofing attack detection, while ensuring optimal performance, resource utilization, and scalability, paving the way for more efficient and adaptable solutions for real-world applications.

Few studies~\cite{al2023unified, watanabe2022spoofing, keresh2024liveness} have explored the use of Transformers for unified facial recognition and physical spoof detection. Mostly, a dual-head Transformer architecture 
that shares a common backbone and adds two classification heads at the end for joint recognition and physical attack detection, %
has been used for the unified task~\cite{al2023unified}.
However, face recognition depends on high-level, identity-specific features, while physical or digital attack detection relies on fine-grained, local details such as texture and noise~\cite{ watanabe2022spoofing, keresh2024liveness}.
Dual-head architectures, which share a common feature space, often underperform due to the mismatch in feature requirements for recognition and spoof classification tasks~\cite{al2023unified}. 
Recent research addresses this limitation by using intermediate features from the Transformer for physical attack detection and identity-specific features from the last layers for face recognition~\cite{al2023unified}. 

With the availability of unified attack datasets comprising both physical and digital facial attacks for the same identities, recently unified physical and digital attack detection methods are proposed using optimized data augmentation~\cite{fang2024unified} and balanced loss functions~\cite{zhang2016joint}, and the joint use of CLIP and a Multi-Layer Perceptron (MLP)~\cite{lin2024robust}.
However, these unified attack detection studies \textit{do not} include facial recognition.

This paper proposes a unified model for face recognition and joint detection of physical and digital spoof attacks for the \textit{first} time.
To achieve this, we leverage the Swin Transformer architecture\cite{liu2021swin} as the backbone along with HiLo attention~\cite{pan2022fast} and a CNN-based classification in local blocks of the Swin Transformer for joint detection of physical and digital spoofing attacks (as a part of the unified attack detection~(UAD) module). 
The global classification head is appended at the end of the Transformer backbone for face representation and matching (FRM).
HiLo attention is used for joint physical-digital attack detection to effectively capture high-frequency details, such as subtle artifacts and fine-grained textures, along with low-frequency information that preserves the broader contextual structure of the image or video~\cite{zakkam2025codeit}.
This combination of localized detailed analysis and global contextual understanding strengthens the model's capability to accurately and reliably detect both kind of spoofing attacks.
By separating attention heads into groups based on high-frequency and low-frequency patterns, HiLo attention enables the model to detect a wide range of attacks, from localized manipulations to global distortions. 
Data enhancement simulating physical and digital cues is added to further facilitate joint physical-digital attack detection.
Figure~\ref{fig:architecture} shows the overall architecture of the proposed unified model for face representation and matching (FRM) and joint physical-digital attack detection (UAD module).

\noindent~\textbf{Contributions:} In summary, the main \textbf{contributions} of this paper are summarized as follows:
\begin{itemize}
    \item Developing a unified model capable of performing both face recognition and joint physical and digital attack detection using the Swin Transformer as a backbone.
    \item Using HiLo attention along with CNN architecture appended to intermediate transformer blocks (UAD module) to capture both high-frequency and low-frequency image features required for joint physical and digital facial attack detection.
    \item Adding data augmentations simulating physical and digital attack cues resulting in UAD module efficacy in detecting both physical-digital as well as unknown attacks. 
    \item Extensive evaluation of the unified model's capability in face recognition, physical, and digital facial spoof attack using diverse datasets and across unknown attack detection scenario.
\end{itemize}

\section{Related Work}
\subsection{Face Representation and Matching}

Face recognition systems have significantly advanced, evolving from traditional approaches such as holistic methods~\cite{belhumeur1997eigenfaces} and handcrafted features~\cite{liu2002gabor, cao2010face} to modern deep learning-based models.
Traditional methods struggled to manage large intra-class variations and inter-class similarity, a limitation effectively addressed by deep-learning based methods~\cite{taigman2014deepface, tan2010face, parkhi2015deep, schroff2015facenet}.
Notable CNN-based systems such as DeepFace~\cite{hsu2024pose}, DeepID~\cite{wu2024quality}, and FaceNet~\cite{ahn2024uncertainty} have demonstrated notable performance in face recognition.

Recent studies demonstrated the efficacy of Vision Transformers for face recognition. 
For instance,~\cite{zhong2021face} utilized Vision Transformer (ViT) models for face matching, demonstrating state-of-the-art performance on large-scale datasets.
~\cite{qin2023swinface} demonstrate the effectiveness of the Swin Transformer in face recognition, facial expression, and age estimation tasks.
Similarly,~\cite{sun2022part} proposed fViT, a Vision Transformer baseline that surpasses state-of-the-art face recognition methods.
They also introduced part fViT, a part-based approach for facial landmarks and patch extraction, obtaining state-of-the-art accuracy on multiple facial recognition benchmarks.~\cite{talemi2024catface} proposed a cross-attribute-guided Transformer framework combined with self-attention distillation to enhance low-quality face recognition.


\subsection{Unified Models}
The study in~\cite{al2023unified} proposed a dual-head approach for unified face recognition and physical attack detection that incorporates separate classification heads for each of these tasks, utilizing shared features from a common Vision Transformer (ViT) backbone. However, the performance of this method was suboptimal due to the differing feature requirements of the two tasks. To address this issue, the authors of~\cite{al2023unified} introduced a unified framework that leverages local features from the intermediate layers of the ViT for detecting physical attacks while utilizing the class token from the final layer of the ViT for face recognition. Similarly,~\cite{suganthi2022deep} proposed a unified approach for both face recognition and physical attack detection by combining the Fisherface algorithm with local binary pattern histograms (LBPH) and deep belief networks. In this method, Fisherface is used to recognize faces by reducing the dimensionality in the facial feature space through LBPH, while a deep belief network with a Restricted Boltzmann Machine serves as a classifier for detecting deepfake attacks.
~\cite{liu2019identity} combines Kinect sensors with FaceNet to enhance liveness detection and identity authentication. 
These aforementioned models perform unified face matching and physical attack detection. In~\textit{contrast}, our proposed unified model performs face matching along with joint physical-digital spoof detection.

\section{Proposed Method}

The primary objective of this work is to develop a multi-task model capable of obtaining optimal performance in both face representation and matching (FRM) and unified attack detection (UAD) (both physical and digital spoofing attacks).
To accomplish this, we use the Swin Transformer as the backbone and evaluate a hybrid multi-task architecture.
The FRM head is placed at the end of the backbone to leverage deep features critical for face recognition (see Figure~\ref{fig:architecture}).
For the UAD module, HiLo attention with a CNN is used to leverage the rich local features extracted from the intermediate layers of the Swin Transformer.
Further, the training data is augmented using Simulated Physical Spoofing Clues (SPSC) and Simulated Digital Spoofing Clues (SDSC) methods (detailed in section~\ref{sim_spoof}), enabling effective training of UAD module for joint physical-digital spoofing attack detection.


\begin{figure}
    \centering
    \includegraphics[width=\columnwidth]{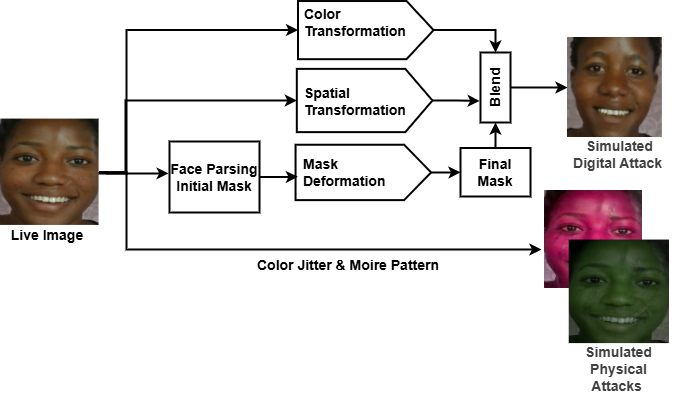} 
    \caption{Illustration of the data augmentation process for a live face sample into physical and digital attack cues using SPSC and SDSC augmentation techniques.}
    \label{fig:simulation}
\end{figure}

\begin{figure*}
    \centering
    \includegraphics[width=.65\textwidth]{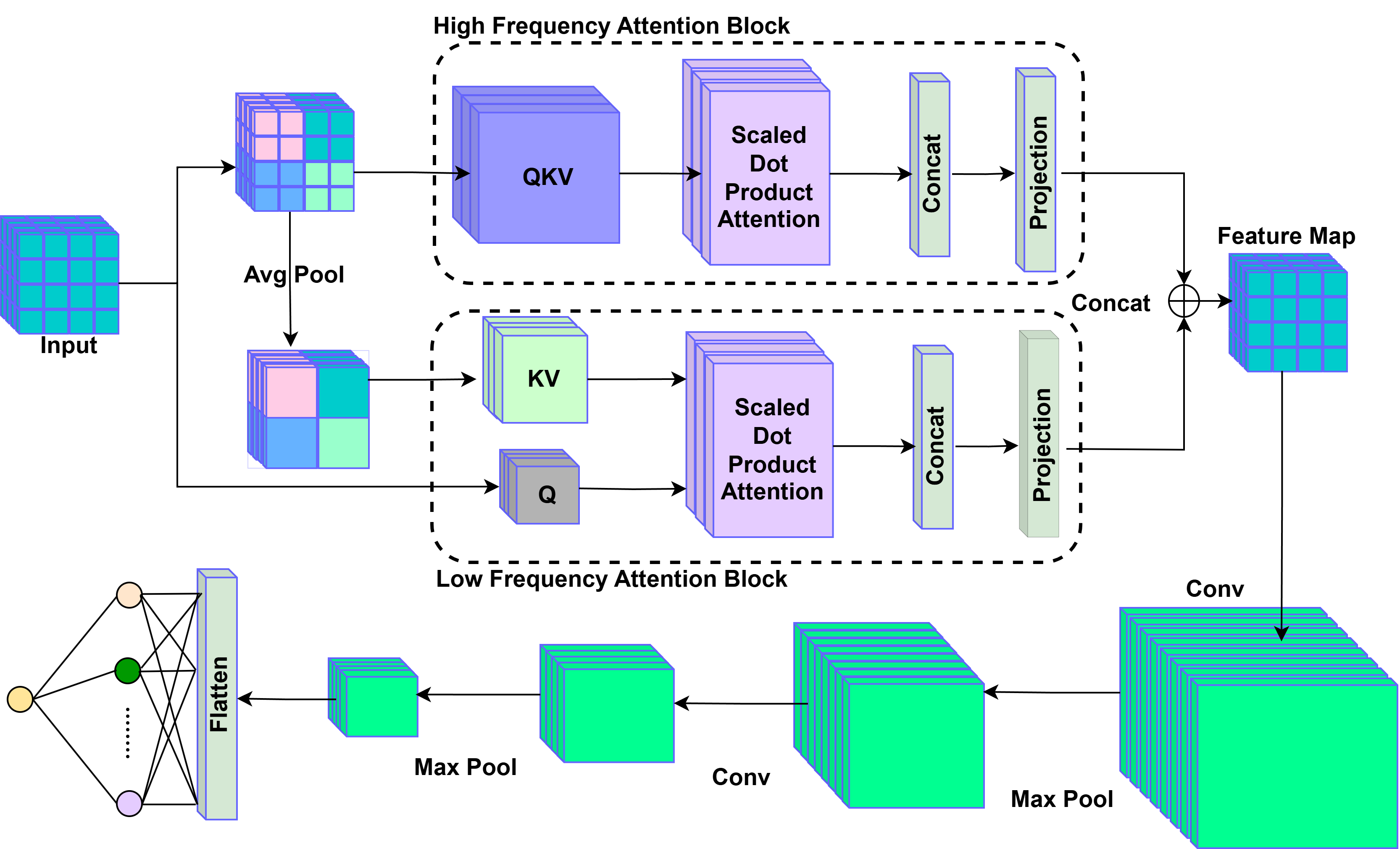}
    \caption{UAD module leveraging Stage $3$ intermediate features along with HiLo attention and a cascaded CNN for joint physical-digital attack detection. High-frequency and low-frequency blocks use an attention module defined as a scaled dot-product function, where query, key, and value matrices are denoted as Q, K, and V, respectively. }
    \label{fig:UAD}
\end{figure*}

\subsection{Swin Backbone}

The base Swin Transformer backbone~\cite{liu2021swin} has four hierarchical stages that reduce spatial dimensions and increase feature depth via patch merging. Each stage includes Swin Transformer blocks with Shifted Window Multi-Head Self-Attention (SW-MHSA), a Feedforward Network (FFN), Layer Normalization (LN), and residual connections.
The input image $x \in \mathbb{R}^{H \times W \times C}$ (where $H$, $W$, and $C$ represent the height, width, and number of channels of the image, respectively) is first divided into non-overlapping patches of size $4 \times 4$ using a patch partitioning layer, resulting in an initial sequence of patch embeddings $Z_0 \in \mathbb{R}^{\frac{H}{4} \times \frac{W}{4} \times C'}$, where $C' = 128$ represents the embedding dimension of each patch for the base model.
This initial patch embedding forms the input to the hierarchical architecture, with each stage processing and refining the features for higher-level representations.

The SW-MHSA mechanism processes patches within local windows of size $M \times M$, where $M=7$ in the base Swin Transformer, reducing computational complexity compared to global attention.
To enable cross-window interaction, the shifted window approach alternates between regular and shifted windows across layers.
Let $Z^\ell = [z^\ell_1, z^\ell_2, \dots, z^\ell_n] \in \mathbb{R}^{n \times C'}$ denote the output of the $\ell$-th Swin Transformer block, where $n$ is the number of patch tokens and $C'$ is the embedding dimension.
The Swin Transformer processes the input image hierarchically through multiple stages, with patch merging layers at each stage reducing the spatial size of the token map while increasing the embedding dimensions.
The hierarchical structure produces a compact representation $Z_L \in \mathbb{R}^{N \times D}$ after the final stage, where $N$ represents the number of tokens in the final stage, determined by the resolution of the input and the down-sampling factor and $D$ is the dimensionality of the output features.

Typically, the final feature map output of the last stage can be pooled or directly used as a latent representation for downstream tasks, allowing the Swin Transformer backbone to act as an encoder $E_\omega$, mapping an input image $x$ to its latent feature representation $Z_L$, i.e., $Z_L = E_\omega(x)$.
This hierarchical architecture allows the Swin Transformer to effectively capture local and global features for classification and detection tasks.

\subsection{Unified Attack Detection Module}

The Unified Attack Detection~(UAD) module combines HiLo attention with a lightweight convolutional network (CNN) to enhance spoofing attack detection by leveraging intermediate features from the Swin Transformer backbone, as these shallow features effectively capture local patterns essential for joint physical-digital spoofing attack detection.
The HiLo attention module~\cite{pan2022fast} processes these features by separating attention into high-frequency (Hi-Fi) and low-frequency (Lo-Fi) components, extracting local details and global patterns separately.
The Hi-Fi path, with $4$ attention heads, uses local window self-attention over  \(2 \times 2\) windows to capture fine-grained details efficiently, while the Lo-Fi path applies average pooling within each window to extract low-frequency signals for modeling relationships with query positions. Features from both paths are concatenated and processed through the CNN, which comprises two convolutional layers, max-pooling, and fully connected layers.
The final output, predicting live vs. spoofed faces, is obtained via a sigmoid activation function, enabling joint physical-digital attack detection, as shown in Figure~\ref{fig:UAD}.
\subsection{Simulating Spoofing Cue Augmentation}
\label{sim_spoof}
The Simulated Physical Spoofing Clues (SPSC) augmentation ~\cite{he2024joint} enhances spoof detection by simulating physical attack characteristics through ColorJitter and moiré pattern augmentation (Figure~\ref{fig:simulation}). ColorJitter replicates color distortions in print attacks by adjusting brightness, contrast, saturation, and hue, while moiré pattern augmentation mimics artifacts from replay attacks through pixel remapping and polar transformations. This strategy diversifies the dataset and improves model robustness against spoofing.

The Simulated Digital Spoofing Clues (SDSC) augmentation technique~\cite{he2024joint} replicates digital forgery artifacts, such as face swapping, through a three-step process. 
First, it duplicates the original image to create pseudo-source and target images, augmented with color (e.g., hue, brightness) and spatial transformations (e.g., resize, translate), producing misaligned boundaries (see Figure~\ref{fig:simulation}). 
Second, a face mask is generated using a face parsing algorithm and deformed with spatial transformations, elastic distortions, and blurring. 
Finally, the pseudo-source and target images are blended using the deformed face mask to create a synthetic forgery, enriching the dataset with realistic artifacts to enhance the model's ability to detect such attacks.
\subsection{Our Unified Architecture: FRM and UAD}

The proposed unified architecture employs a Swin Transformer backbone to perform both FRM and UAD tasks, as depicted in Figure~\ref{fig:architecture}.
Initially, the backbone is trained for FRM, utilizing global and deep features (\(7 \times 7 \times 1024\)) extracted from its final stage.
This is facilitated through appending an MLP-based FRM head at the 
end of the backbone for face recognition. 
This head utilizes the output features of the Swin Transformer and applies L2 normalization to the feature embeddings, mapping them onto a unit hypersphere.
This hyperspherical normalization ensures consistent embedding magnitudes, allowing the ArcFace loss function (see equation~\ref{eq:frm_loss}) to effectively enhance class separability and ensure robust performance.

Given $N_{id}$ training samples and $C$ unique identities in the dataset, the loss function is defined as
\begin{equation}
\mathcal{L}_{FRM} = -\frac{1}{N_{id}} \sum_{i=1}^{N_{id}} \log \frac{e^{s \cos \tilde{\theta}_{y_i^{id}}}}{e^{s \cos \tilde{\theta}_{y_i^{id}}} + \sum_{j=1, j \neq y_i^{id}}^{C} e^{s \cos \theta_j}},
\label{eq:frm_loss}
\end{equation}
where $\cos \tilde{\theta}_{y_i^{id}} = \cos(\theta_{y_i^{id}} + m)$. 
Here, $\cos \theta_r$ represents the cosine similarity between the normalized feature vector $v$ and the weight vector $W_r$, with $\|W_r\|$ and $\|v\|$ denoting their $L_2$ norms. 
The parameters $s$ (scale) and $m$ (angular margin) control the strength of the feature scaling and margin, respectively. 
By modifying the decision boundaries in the angular space, the ArcFace loss significantly improves the network's ability to distinguish between different identities, making it ideal for face recognition tasks~\cite{deng2019arcface}.

After FRM training, the backbone is frozen, and the UAD head is appended at the intermediate layer of Stage $3$ (\(14 \times 14 \times 512\)), where shallow features effectively capture local and attack-specific patterns.
This stage is preferred because it contains $18$ attention blocks, significantly more than the 2 blocks in other stages.
These multiple blocks represent varying levels of feature abstraction, providing a broader range of feature representations to test and determine the optimal level for the detection task.
The UAD module also utilizes a HiLo Attention module to extract both high- and low-frequency features from the intermediate outputs of Stage $3$, followed by a lightweight convolutional network for binary classification, trained using Binary Cross-Entropy loss (see equation~\ref{eq:UAD_loss}).
\begin{equation}
L_{\text{SD}} = -\frac{1}{N_{\text{s}}} \sum_{i=1}^{N_{\text{s}}} \left[ y_i^{\text{s}} \log(\hat{y}_i^{\text{s}}) + (1 - y_i^{\text{s}}) \log(1 - \hat{y}_i^{\text{s}}) \right],
\label{eq:UAD_loss}
\end{equation}
where $N_{\text{s}}$ is the number of attack samples (physical and digital), $y_i^{\text{s}} \in \{0, 1\}$  is the ground truth label for sample $i$, and $\hat{y}_i^{\text{s}}$ is the predicted probability for the same sample.
Here, the labels $0$ and $1$ represent the spoof attack (both digital and physical) and bonafide face image, respectively. 
Training of UAD for attack detection is enhanced using SPSC and SDSC augmentations, simulating print, replay, and digital attacks.

Additional training configurations are also explored, where the UAD  head is appended at the final layer of the backbone and trained with and without freezing the backbone, to evaluate the performance trade-off for unified FRM and UAD tasks as shown in Figure~\ref{fig:architecture}.
This comprehensive training strategy allows for assessing the effectiveness of intermediate and final layer features for multi-tasking performance.
\section{Experimental Implementations}

\subsection{Datasets}
The CASIA-WebFace dataset~\cite{yi2014learning} is utilized for training the face recognition model.
The FaceForensics++~\cite{rossler2019faceforensics++} and CelebDF~\cite{li2020celeb} datasets are utilized for benchmarking the accuracy of the face recognition model across datasets.
Similarly, the UniAttack dataset~\cite{fang2024unified}, augmented using the SPSC and SDSC methods, is used to train the UAD module.
FaceForensics++(FF++), SIW-Mv2~\cite{guo2022multi} and MSU Mobile Face Spoofing Database (MSU-MFSD)~\cite{wen2015face} datasets are used for the evaluation of the UAD module.
Further, SiW-Mv2 and Diverse Fake Face Dataset~(DFFD)~\cite{dang2020detection} dataset are used to test the model performance on unknown physical and digital attacks, respectively.
Face crops from facial images extracted from the videos were obtained using the Multi-task Cascaded Convolutional Neural Network~(MTCNN)\cite{zhang2016joint}, a deep learning framework designed for efficient face detection and alignment.
The cropped facial images were resized to $224\times 224$ pixels for further processing.

To evaluate the accuracy of the FRM trained on the CASIA-WebFace dataset, $10,000$ image pairs with matching identities and $10,000$ pairs with non-matching identities were selected from the live image sets of FaceForensics++ and CelebDF.
The Swin Transformer model was employed for feature extraction, and cosine similarity was used to measure the model's performance.
The UAD model was trained using the UniAttack dataset, augmented with SPSC and SDSC techniques.
For cross-dataset evaluation, $20,000$ live and $20,000$ spoof images (FaceShift, FaceSwap, and Face2Face) were selected from FaceForensics++, $5,000$ live and $5,000$ spoof images (replay and print attacks) from the SiW-Mv2 dataset, and $175$ live and $525$ spoof images from the MSU-MFSD dataset.

The model's performance against unknown attacks was evaluated using a variety of spoofing categories from the SiW-Mv2 and DFFD datasets.
Physical spoof categories included Makeup Cosmetics, Makeup Impersonation, Makeup Obfuscation, Mannequin, Mask Half Mask, Mask Paper Mask, Mask Transparent Mask, as well as partial spoofs such as Partial Eye, Partial Funny Eye Glasses, Partial Mouth, and Partial Paper Glasses, along with Silicone masks.
The DFFD dataset comprises of a rich variety of manipulations, including identity and expression swaps sourced from FaceForensics++, facial attribute edits using FaceAPP, and synthetic faces created with GANs like StyleGAN and PGGAN.
It offers high-quality images that represent a broad spectrum of manipulations, from subtle attribute changes to fully synthetic faces.

\subsection{Implementation Details}

The Swin base model, pretrained on ImageNet-21K is used as the backbone.
It consists of four stages with $2$, $2$, $18$, and $2$ transformer blocks in each stage, respectively.
The model processes images by dividing them into patches and progressively reducing spatial dimensions while increasing feature depth across stages~\cite{liu2021swin}.
The FRM head includes a fully connected layer that projects the $d$-dimensional output of the Swin Transformer backbone into a $1024$-dimensional face embedding.
The feature embedding is normalized and ArcFace loss is applied with an angular margin to enforce inter-class separability and intra-class compactness, ensuring highly discriminative features optimized for face recognition.
This FRM head is fine-tuned on the CASIA-WebFace dataset using the ArcFace loss (scale $s=32$, margin $m=0.5$) and optimized with the stochastic gradient descent (SGD) algorithm at a fixed learning rate of $10^{-3}$.
Training is performed with a batch size of $64$ using an early stopping mechanism to avoid overfitting. 

The total size of our FRM system model, which combines the Swin Transformer backbone with $86.7$M parameters and the ArcFace loss function with $10.8$M parameters (for 10,572 classes), is $97.5$M parameters.

Similarly, the UAD module comprising of an immediate HiLo attention block, followed by two convolutional layers ($64$ and $32$ layers with $3\times 3$ kernel), a fully connected layer of size $128$, and a sigmoid activation is appended to each intermediate block of Stage $3$, as well as to the final layer, resulting in $19$ independent classifiers ($18$ classifiers from Stage $3$).
For each UAD module training, the UniAttack dataset and its augmentations (SPSC and SDSC) are utilized. It is trained under two settings: with the Swin Transformer backbone frozen and unfrozen.
All UAD modules are trained under identical hyperparameter settings for consistent comparison.
The UAD model's architecture is lightweight, comprising approximately $1.2$ million parameters, facilitating efficient training and inference for joint spoofing attacks.
Trained using binary cross-entropy loss, the classifiers' parameters are optimized with SGD at a learning rate of $10^{-3}$ using a batch size of $64$ using an early stopping mechanism.

\subsection{Evaluation Metrics}
The FRM task is assessed by measuring face-matching accuracy, determined through face verification for fair comparison with~\cite{al2023unified}.
The performance of the UAD classifier is evaluated using metrics such as Attack Presentation Classification Error Rate (APCER), Bona Fide Presentation Classification Error Rate (BPCER), and classification accuracy~\cite{fang2024unified,zhang2016joint}. 
Additionally, the Equal Error Rate (EER) is computed to provide a more precise and comprehensive evaluation of both the FRM and UAD tasks.

\section{Results and Discussion}

\noindent \textbf{FRM Evaluation.} Firstly, the Swin Transformer backbone pretrained on ImageNet-$21$K, combined with the FRM head, is fine-tuned on the CASIA-WebFace dataset and evaluated on FF++ and CelebDF datasets for face recognition tasks.
Our model obtained $99.43\%$ accuracy on FF++, surpassing the $98.96\%$ reported in~\cite{al2023unified} and is at par with $99.53\%$ across datasets in~\cite{deng2019arcface}, also shown in Table~\ref{tab:Comparision FRM}.

 The same model is fine-tuned for the UAD head appended at the end using the augmented UniAttack dataset and evaluated on the FF++ and CelebDF datasets for FRM (Table ~\ref{tab:FRM}). Results show that fine-tuning the Swin Transformer for UAD head reduces FRM accuracy, with a $4.4$\% drop on FF++ and $8.73$\% on CelebDF, highlighting the complimentary nature of features used for UAD and FRM tasks.


 \begin{table}
  \centering
  \resizebox{0.48\textwidth}{!}{%
  {\small{
  \begin{tabular}{@{}lcc|cc@{}}
    \toprule
    \multirow{3}{*}{\shortstack{Training \\ Scenario}} & \multicolumn{2}{c|}{\shortstack{FF++}} & \multicolumn{2}{c}{\shortstack{Celeb DF}} \\
    \cmidrule(lr){2-3} \cmidrule(lr){4-5}
    & Acc (\%) & EER (\%) & Acc (\%) & EER (\%) \\
    \midrule
    FRM Finetuned & \textbf{99.43} & \textbf{1.23} & \textbf{95.23} & \textbf{4.00} \\
    UAD Finetuned & 95.03 & 5.46 & 86.50 & 15.3 \\
    \bottomrule
  \end{tabular}
  }}
  }
  \caption{Face matching accuracy on FF++ and CelebDF datasets: (Row 1) Swin Transformer model fine-tuned for FRM using CASIA-WebFace dataset and (Row 2) the same model further fine-tuned for UAD using augmented UniAttack dataset, tested for FRAM (face recognition performance).}
  \label{tab:FRM}
\end{table}

\begin{table}
  \centering
  {\small{
\begin{tabular}{@{}c|c|c@{}}
    \toprule
     Study & Model & Accuracy (\%)\\
    \midrule
    Deng et al.~(2022) \cite{deng2019arcface} & ResNet50  & 99.53 \\
    Al-Refai et al.~(2023) \cite{al2023unified} & ViT & 98.96  \\
    \textbf{Ours (FRM Finetuned)} & \textbf{Swin-T} &\textbf{ 99.43}   \\
    \bottomrule
  \end{tabular}
  }}
  \caption{Performance of existing and our proposed model for FRM task.}
  \label{tab:Comparision FRM}
\end{table}

\noindent \textbf{UAD Evaluation.} UAD was evaluated under two scenarios.
In the first scenario, the UAD head was appended to the end of the Swin transformer, and the model was fine-tuned with and without freezing the backbone.
In both setups, the UAD performance remains suboptimal, indicating that the features extracted at deeper layers are not well-suited for UAD tasks as illustrated in Table~\ref{tab:UAD_end}.
This is likely because deeper layers focus on high-level features, while UAD relies more on low-level features essential for effective attack detection.


\begin{table}
  \centering
  \resizebox{0.48\textwidth}{!}{%
  {\large{
  \begin{tabular}{@{}c|cc|cc|cc@{}}
    \toprule
    \multirow{3}{*}{\shortstack{Finetuned \\ (UAD Head)}} 
    & \multicolumn{2}{c|}{\shortstack{FF++}} 
    & \multicolumn{2}{c|}{\shortstack{SiW}} 
    & \multicolumn{2}{c}{\shortstack{MSU-MFSD}} \\
    \cmidrule(lr){2-3} \cmidrule(lr){4-5} \cmidrule(lr){6-7}
    & Acc (\%) & EER (\%) & Acc (\%) & EER (\%) & Acc (\%) & EER (\%) \\
    \midrule
    \shortstack{ SWIN unfrozen} & 51.32 & 44.78 & 55.34 & 49.46 & 75.65 & 58.53 \\
    \midrule
    \shortstack{SWIN frozen} & \textbf{74.67} & \textbf{26.92} & \textbf{63.45} & \textbf{38.47} & \textbf{75.54} & \textbf{51.17} \\
    \bottomrule
  \end{tabular}
  }}
  }
  \caption{Cross-dataset performance comparison for UAD head fine-tuned with and without freezing Swin transformer backbone. Training was performed on the augmented UniAttack dataset. Results include accuracy (Acc) and equal error rate (EER) for FF++ (Digital: Deepfakes, Face2face and Faceswap), SiW (Physical: Paper and Replay), and MSU-MFSD~(Physical: Paper and Replay) datasets.}
  \label{tab:UAD_end}
\end{table}

In the second scenario, the Swin Transformer backbone was frozen, and UAD modules were appended at all intermediate blocks of Stage $3$ of the Swin Transformer, with each UAD module trained independently.
Performance was evaluated on FF++, SiW-Mv2, and MSU-MFSD datasets, trained using the augmented UniAttack dataset , as shown in Table~\ref{tab:stage3_blocks}. FF++ includes attack types such as digital Deepfake, Face2Face, and FaceSwap, while SiW-Mv2 and MSU-MFSD includes Physical Replay and Print attacks. 
Among all intermediate blocks, the sixth block of Stage $3$ (i.e., Stage 3 Block $5$) demonstrated the best performance across the evaluated datasets, making it most suitable for UAD tasks.
This block extracts features that effectively balance low-frequency abstract representations with high-frequency texture-level details, which is crucial for joint attack detection, as it requires both broader structural inconsistencies and subtle local artifacts, which are key indicators of spoofing attempts. Therefore the results from the UAD module appended at the sixth block of Stage $3$ (i.e., Stage 3 Block $5$) are used as performance metrics for further analysis in this study. Additionally, since the backbone weights remain unchanged, the performance of FRM remain unaffected. 

For digital spoof detection, our model achieved accuracies of $97.2\%$ on the FF++, surpassing the performance over state-of-the-art deepfake detectors listed in Table~\ref{tab:Comparision Deepfake Detection}.
Also, for physical spoof detection, our model achieved $86.8\%$ accuracy on SiW-Mv2, surpassing most state-of-the-art physical spoof detectors as illustrated in Table ~\ref{tab:Comparision PA Detection}. These results confirms \textit{equivalent} performance of our unified model for physical and digital attack detection over SOTA physical and digital (deepfake) attack detector baselines. 
The results highlights that UAD performs better when appended to intermediate blocks, as these blocks focus more on low-level features essential for joint physical-digital attack , over UAD head appended at the end of the Swin Transformer backbone.

\begin{table}
  \centering
  \resizebox{0.48\textwidth}{!}{%
  {\small{
  \begin{tabular}{@{}c|cc|cc|cc@{}}
    \toprule
    \multirow{2}{*}{\shortstack{Stage 3 \\ Blocks}} 
    & \multicolumn{2}{c|}{FF++} 
    & \multicolumn{2}{c|}{SiW-Mv2} 
    & \multicolumn{2}{c}{MSU-MFSD} \\
    \cmidrule(lr){2-3} \cmidrule(lr){4-5} \cmidrule(lr){6-7}
    & Acc (\%) & EER (\%) & Acc (\%) & EER (\%) & Acc (\%) & EER (\%) \\
    \midrule
    0 & 95.4 & 5.2 & 67.4 & 33.5 & 75.3 & 30.6 \\
    1 & 96.5 & 4.4 & 69.7 & 32.3 & 74.7 & 30.3 \\
    2 & 96.1 & 4.1 & 74.0 & 29.3 & 76.7 & 32.7 \\
    3 & 97.0 & 3.9 & 76.6 & 25.2 & 77.1 & 28.4 \\
    4 & 96.6 & 4.3 & 81.5 & 19.7 & 75.0 & 23.5 \\
    \textbf{5} & \textbf{97.2} & \textbf{4.1} & \textbf{86.8} & \textbf{14.2} & \textbf{79.1} & \textbf{19.5} \\
    6 & 96.3 & 4.1 & 82.3 & 20.6 & 75.8 & 31.3 \\
    7 & 95.1 & 5.3 & 83.4 & 20.2 & 76.9 & 29.1 \\
    8 & 95.2 & 5.8 & 81.8 & 21.1 & 81.4 & 22.7 \\
    9 & 93.3 & 7.6 & 77.2 & 23.4 & 78.0 & 22.4 \\
    10 & 91.9 & 9.1 & 78.3 & 23.4 & 76.4 & 27.7 \\
    11 & 91.8 & 9.3 & 75.7 & 25.6 & 75.5 & 26.1 \\
    12 & 91.0 & 9.7 & 75.3 & 25.6 & 75.7 & 29.9 \\
    13 & 90.5 & 10.2 & 72.3 & 28.6 & 75.4 & 38.2 \\
    14 & 89.6 & 11.7 & 73.4 & 26.3 & 75.7 & 37.4 \\
    15 & 88.7 & 12.5 & 74.7 & 25.3 & 75.3 & 40.2 \\
    16 & 86.1 & 14.3 & 64.3 & 37.4 & 75.7 & 42.4 \\
    17 & 86.3 & 16.3 & 66.3 & 34.7 & 75.2 & 40.2 \\
    \bottomrule
  \end{tabular}
  }}
  }
  \caption{Performance comparison of UAD modules appended to different  Stage 3 blocks. Results include accuracy (Acc) and equal error rate (EER) for FF++~(digital: deepfake, face2face and faceswap), SiW-Mv2~(physical: replay and print), and MSU-MFSD~(physical: replay and print) datasets.}
  \label{tab:stage3_blocks}
\end{table}


\begin{table}
  \centering
  \resizebox{0.48\textwidth}{!}{%
  {\small{
\begin{tabular}{@{}c|c|c@{}}
    \toprule
     Study & Model & Accuracy (\%) \\
    \midrule
    Chollet et al.~(2017) \cite{chollet2017xception} & Xception  & 74.1 \\
    Liu et al.~(2021) \cite{liu2021proceedings} & Seferbekov  & 73.5  \\
    Wang et al.~(2019) \cite{Bittner_2019_CVPR_Workshops} & Eff.B1 + LSTM & 68.3   \\
    Cozzolino et al.~(2021) \cite{cozzolino2021id} & ID-Reveal & 81.7 \\
    Zakkam et al.~(2025) \cite{zakkam2025codeit} & CoDeiT-XL  & 88.1 \\
    \textbf{Ours~(Unified Approach)} & \textbf{UAD} & \textbf{ 97.2}  \\
    \bottomrule
  \end{tabular}
  }}
  }
  \caption{Comparing performance of our model with state-of-the-art deepfake detectors  evaluated on the FF++ dataset.}
  \label{tab:Comparision Deepfake Detection}
\end{table}

\begin{table}
  \centering
  \resizebox{0.48\textwidth}{!}{%
  {\small{
\begin{tabular}{@{}c|c|c@{}}
    \toprule
     Study & Model & Accuracy (\%)\\
    \midrule
    Arora et al. (2021) \cite{arora2022robust} & Autoencoders  & 60.11 \\
    
    Deb and Jain (2021) \cite{deb2020look} & SSR-FCN  & 80.1   \\
    
    Niraj et al. (2023) \cite{thapa2023presentation} & Mobilenet/PAD-CNN & 92.0/89.0   \\
    
    Yu et al. (2020) \cite{yu2020searching} & CDCN & 81.7   \\
    
    
    \textbf{Ours~(Unified Approach)} & \textbf{UAD} & \textbf{ 86.8}    \\
    \bottomrule
  \end{tabular}
  }}
  }
  \caption{Comparison of our model's performance with state-of-the-art physical spoof detectors.} 
  \label{tab:Comparision PA Detection}
\end{table}

Additionaly, we evaluated our unified model on unknown physical and digital spoofing attacks from the SiW-Mv2 and DFFD datasets, respectively.
Table~\ref{tab:UAD_unknown_physical} and Table~\ref{tab:UAD_unknown_digital} demonstrate the results of our UAD model on unknown attacks, with the UAD module appended at \text{5\textsuperscript{th}} block of Stage $3$.
As can be seen from Table~\ref{tab:UAD_unknown_physical}, the UAD module also obtains acceptable performance on unknown physical attacks with an average accuracy of $80.11\%$, with the highest being $98.12\%$ for Mask Papermask and the lowest being $59.06\%$ for Partial Paperglasses. The high accuracy of Mask Papermask is due to its distinct texture and structural artifacts, such as sharp edges and material-specific features, which are easily detected.
In contrast, the lower accuracy for Partial Paperglasses stems from its minimal and localized changes, often confined to small regions like the eyes or glasses, making detection more challenging.
Notably, the performance of our UAD on unknown physical attacks exceeds that reported in~\cite{al2023unified} for all types of Makeup attacks. While the accuracy for Makeup spoofing in~\cite{al2023unified} was $61.73\%$, our model achieves high accuracy of $74.23\%$, $72.45\%$, and $89.73\%$ for Makeup Cosmetics, Makeup Impersonation, and Makeup Obfuscation spoof types, respectively.
Furthermore, the accuracy of our unified model on unknown attacks namely Silicone, Partial Mouth, Partial Eye, and Mask Halfmask—categories not addressed in~\cite{al2023unified}—are $93.56\%$, $90.05\%$, $81.17\%$, and $80.91\%$, respectively.

As shown in Table~\ref{tab:UAD_unknown_digital}, our UAD module achieves an average accuracy of \textbf{72.36\%} on unknown digital attacks which are attribute manipulation based deepfakes from DFFD~\cite{dang2020detection} dataset. Notably, the performance of our model is at par with the results reported in~\cite{nadimpalli2024proactive,10.1145/3700146} on unknown digital attacks, highlighting its effectiveness to unknown digital attacks.




These experiments highlight the efficacy of our proposed unified model in obtaining performance at par with the SOTA individual face recognition and physical and digital spoof detectors across datasets and unknown attack types.

\begin{table}
  \centering
  {\small{
  \begin{tabular}{@{}c|l|c@{}}
    \toprule
    SN & Spoof Type & Accuracy (\%) \\
    \midrule
    1 & Makeup Cosmetics & 74.23 \\
    2 & Makeup Impersonation & 72.45 \\
    3 & Makeup Obfuscation & 89.73 \\
    4 & Mannequin & 87.43 \\
    5 & Mask Halfmask & 80.91 \\
    6 & \textbf{Mask Papermask} & \textbf{98.12} \\
    7 & Mask Transparent Mask & 73.76 \\
    8 & Partial Eye & 81.17 \\
    9 & Partial Funnyeyeglasses & 60.91 \\
    10 & Partial Mouth & 90.05 \\
    11 & Partial Paperglasses & 59.06 \\
    12 & Silicone & 93.56 \\
    \bottomrule
  \end{tabular}
  }}
  \caption{Performance of UAD module appended at an intermediate block (Stage $3$, Block $5$) of the Swin Transformer backbone on unknown physical attacks from the SiW-Mv2 dataset.}
  \label{tab:UAD_unknown_physical}
\end{table}

\begin{table}
  \centering
  {\small{
  \begin{tabular}{@{}c|l|c@{}}
    \toprule
    SN & Spoof Type & Accuracy (\%) \\
    \midrule
    1 & Faceapp & 69.93 \\
    2 & PGGAN\_V1 & 74.70 \\
    3 & PGGAN\_V2 & 72.08 \\
    4 & \textbf{StarGAN} & \textbf{77.34} \\
    5 & StyleGAN\_CelebA & 70.89 \\
    6 & StyleGAN\_FFHQ & 69.27 \\
    \bottomrule
  \end{tabular}
  }}
  \caption{Performance of UAD module appended at an intermediate block ~(Stage $3$, Block $5$) of the Swin Transformer backbone on unknown digital attacks based on attribute manipulation from the DFFD dataset.}
  \label{tab:UAD_unknown_digital}
\end{table}


\section{Conclusion and Future Works}

We proposed a unified model to jointly perform the physical-digital face attack detection and face-matching tasks. 
We exploit Swin Transformer architecture and leverage the local features extracted from the intermediate blocks in conjunction with HiLo attention and CNN for joint spoof detection while using the global features learned by the final layer for face matching.
Experiments conducted in various settings demonstrate that our proposed unified model can achieve equivalent performance for both physical-digital attack detection and face-matching tasks in comparison to SOTA models across datasets and attack types.
As a part of future work, we will evaluate the efficacy of our proposed unified model across different biometrics modalities and novel attack types based on different generation techniques.


\balance
{\small
\bibliographystyle{ieee_fullname}
\bibliography{egbib}
}

\end{document}